# PoseViNet: Distracted Driver Action Recognition Framework Using Multi-View Pose Estimation and Vision Transformer

Neha Sengar, Indra Kumari, Jihui Lee, Dongsoo Har, *Senior Member, IEEE*

*Abstract*—Driver distraction is a principal cause of traffic accidents. In a study conducted by the National Highway Traffic Safety Administration, engaging in activities such as interacting with in-car menus, consuming food or beverages, or engaging in telephonic conversations while operating a vehicle can be significant sources of driver distraction. From this viewpoint, this paper introduces a novel method for detection of driver distraction using multi-view driver action images. The proposed method is a vision transformer-based framework with pose estimation and action inference, namely PoseViNet. The motivation for adding posture information is to enable the transformer to focus more on key features. As a result, the framework is more adept at identifying critical actions. The proposed framework is compared with various state-of-the-art models using SFD3 dataset representing 10 behaviors of drivers. It is found from the comparison that the PoseViNet outperforms these models. The proposed framework is also evaluated with the SynDD1 dataset representing 16 behaviors of driver. As a result, the PoseViNet achieves 97.55% validation accuracy and 90.92% testing accuracy with the challenging dataset.

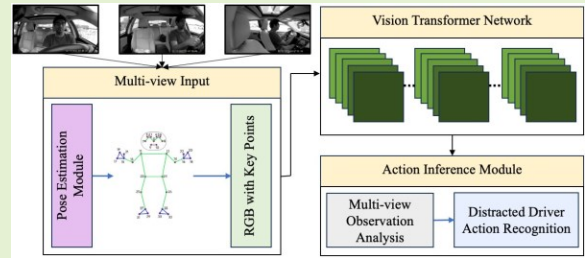

## I. INTRODUCTION

ACCORDING to the Global State of Road Safety Report by the World Health Organization (WHO) [1], 50 million people are wounded, and 1.35 million die in traffic accidents yearly. The National Highway Traffic Safety Administration of the United States (NHTSA) conducted investigations that revealed driver distraction causes a significant number of fatalities and financial losses [2]–[4]. Statistics indicate that driver distraction was a factor in nearly 8.7 % of all accidents in 2019 [5].

Any action that leads the driver to shift their focus from the roadway, such as texting, calling on the phone, eating, drinking, yawning, or conversing with passengers, is seen as a driver distraction [6]. In this paper, we consider driver distraction in the form of adjusting the control panel, drinking, eating, using a phone in the form of texting or calling, reaching behind the driver seat, fixing hair and doing makeup, yawning, picking up from the floor, hand on head, and talking to a fellow passenger for a longer duration. However, it has been analyzed that accidents can be averted if drivers are notified of driver distraction in a timely manner. Use of internet-of-things sensors [7], [8] with simple decoding [9] can help timely notification for preventing accident. Next-generation automobiles are likely to come with advanced driver assistance systems (ADAS) to help in such situations [10].

The ADAS analyzes sensor data obtained by cameras, sonar, lidar, radar, and GPS to identify probable accident scenarios. Notably, the camera has been integrated into the ADAS, which has created new opportunities for computer vision research. Researchers have primarily concentrated on using deep learning (DL) models to solve the driver distraction problem over the past ten years. The DL models have marked numerous improvements in various domains, including object detection [11], [12], safety enhancement [13], and action recognition [14]. These models have also been expanded to include the ability to detect the driver's emotional state [15]–[17]. However, the drawback of the DL models is the lengthy training process required. As a result, these models need a large volume of training data. Additionally, the pooling procedure costs CNN precious information because it leads to

Neha Sengar is with the Robotics Program, Korea Advanced Institute of Science and Technology, Daejeon 34141, South Korea (e-mail: nehass2911@gmail.com).

Indra Kumari is with Korea Institute of Science and Technology Information, Daejeon 34141, South Korea (e-mail: kumariindra7@gmail.com).

Jihui Lee is with the Division of Future Vehicle, Korea Advanced Institute of Science and Technology, Daejeon 34141, South Korea (e-mail: jihui@kaist.ac.kr).

Dongsoo Har is with the CCS Graduate School of Green Transportation, Korea Advanced Institute of Science and Technology, Daejeon 34141, South Korea (e-mail: dshar@kaist.ac.kr).

This work is supported by the Institute for Information communications Technology Promotion (IITP) grant funded by the Korean government (MSIT)(No.2020-0-00440, Development of Artificial Intelligence Technology that continuously improves itself as the situation changes in the real world).

the loss of valuable information that the network has learned from the input data.

The computer vision research area has experienced exponential growth due to the recent development of AI-enabled image processing, where transformer-based approaches are primarily used due to the advent of vision transformer (ViT) [18]. The ViT has outperformed several state-of-the-art techniques for several computer vision tasks such as image classification [19], object identification [20], image colorization [21], semantic segmentation [22], video comprehension [23], low-density vision [24], and more due to its superior generalization ability. These significant qualities of the ViT have stimulated considerable interest in a range of applications of intelligent transportation systems [25], [26].

However, the ViT processes images as a sequence of patches, neglecting the spatial relationships between pixels. To alleviate this, this paper presents a new ViT incorporated with pose estimation, which can provide information about the position and orientation of objects within the image. Incorporating pose estimation into the ViT pipeline helps the PoseViNet focus on significant portions of the image and better understand the spatial context and leverage this information for improved accuracy.

Contributions of this work can be summarized as following:

- A novel framework for the driver distraction recognition, PoseViNet, is proposed. The framework takes driver's poses in multi-view images, unlike one view images taken by current models for driver distraction recognition, to classify the driver's actions into different categories.
- Structural details of the PoseViNet are presented. The PoseViNet consists of pose estimation module, ViT network, and action inference module. The pose estimation module deals with pose estimation, and the ViT network comprising self-attention, multi-head self-attention, and multilayer perception with dropouts, and fully-connected layers recognizes driver distraction, and the action inference module enhances the precision of action recognition.
- Posture information obtained from pose estimation is utilized to provide the most discriminating feature map to the ViT network. The inclusion of posture information focuses on extracting the relevant features from the vital portion of the image instead of the overall image.
- Benchmark testing is performed with the SFD3 dataset [63] representing 10 behaviors of driver. The findings from the comparison with the state-of-the-art models using the SFD3 dataset show that the PoseViNet outperforms these baseline models. With more comprehensive dataset SynDD1 [61] representing 16 behaviors of driver, the PoseViNet is still able to achieve high performance.

The rest of the paper is organized as follows. Section II discusses the related works. The proposed method is explained in Section III using the PoseViNet framework. Section IV presents the experiment settings. Section V shows the result analysis and comparison with state-of-the-art techniques. Section VI concludes the paper with suggestions for future works.

## II. RELATED WORKS

Driver action recognition has emerged as a promising research area of intelligent transportation system. Some researchers have applied traditional machine learning methods for driver behavior recognition. The advancement of computer vision and deep learning techniques have paved the way for significant progress in driver behavior detection. The success of pose estimation opens a new area to explore in driver-distracted behavior recognition [27].

### A. Traditional Methods

Manual feature extraction and machine learning classifiers are the main components of traditional methods. Zhang et al. [28] developed a hidden conditional random field to identify a cell phone after extracting the driver's hands, face, and lips features. Seshadri et al. [29] observed the driver's facial histogram of oriented gradient characteristics and used AdaBoost to track phone usage with an accuracy of 93.9%. Craye and Karray [30] examined the driver's eyes, head, arms, and face in RGB-depth data using AdaBoost and a hidden markov model. Zhao et al. [31] used a tiny contour transform to extract features, and then Random Forest, K-Nearest Neighbour, Multi-layer Perceptron, and Support Vector Machine (SVM) were used to identify distraction, respectively, where SVM achieved an accuracy of 90.63%. Gupta et al. [32] established a system to identify drivers' yawning and sleeping states. The findings in the mentioned literature are all based on small datasets. These machine learning and manual feature-based approaches perform poorly on complicated and noisy datasets and real-time natural driving states are incomprehensible to them.

### B. Deep Learning Models

Deep convolutional networks (DCNNs) are capable of capturing important information in complex data. Yan et al. [33] achieved 99.78% accuracy in driver posture categorization using a DCNN with three convolutional layers, pooling layers, and three fully connected layers. Koesdwiady et al. [34] achieved 95% accuracy using a pre-trained VGG19 [35] model on their dataset. Hssayeni et al. [36] found that ResNet [37] and VGG16 [35] outperform manual feature extraction. With the use of pre-trained AlexNet [38], Inception-V3 [39], and a genetic weighting approach, Abouelnaga et al. [40] retrieved hand and facial characteristics with a precision of 94.29%. They also enhanced the driver's skin segmentation for 90% accuracy using ensemble technique [41]. The ResNet-50 [37], Xception [42], InceptionV3 [39], and VGG19 [35] were used to achieve 97% accuracy by Dhakate and Dash [43]. Initial module, residual module, and hierarchical recurrent neural network combinations were considered for classification by Alotaibi & Alotaibi [44]. Mittal and Verma [45] designed a convolution-based capsule network with attention mechanisms for detecting driving behaviors. Liu et al. [46] developed a teacher and student network for a lightweight driver distraction recognition system. Aljohani [47] developed a framework where CNN models like VGG19 [35], ResNet50 [37], and DenseNet121's [48] feature extractor are initially selected using evolutionary algorithms. Then, two layers of the deep learning model are added as a classification network achieving 99.80% accuracy. Ma and Wang [49] introduced a semi-supervised technique based on the ViT for driver distraction and emotion identification. They created a multi-modal vision transformer, ViTDD, leveraging training signals. While these works initially focused on single-view data, multi-view approaches enhance deep learning robustness by mitigating noise and errors in individual views.

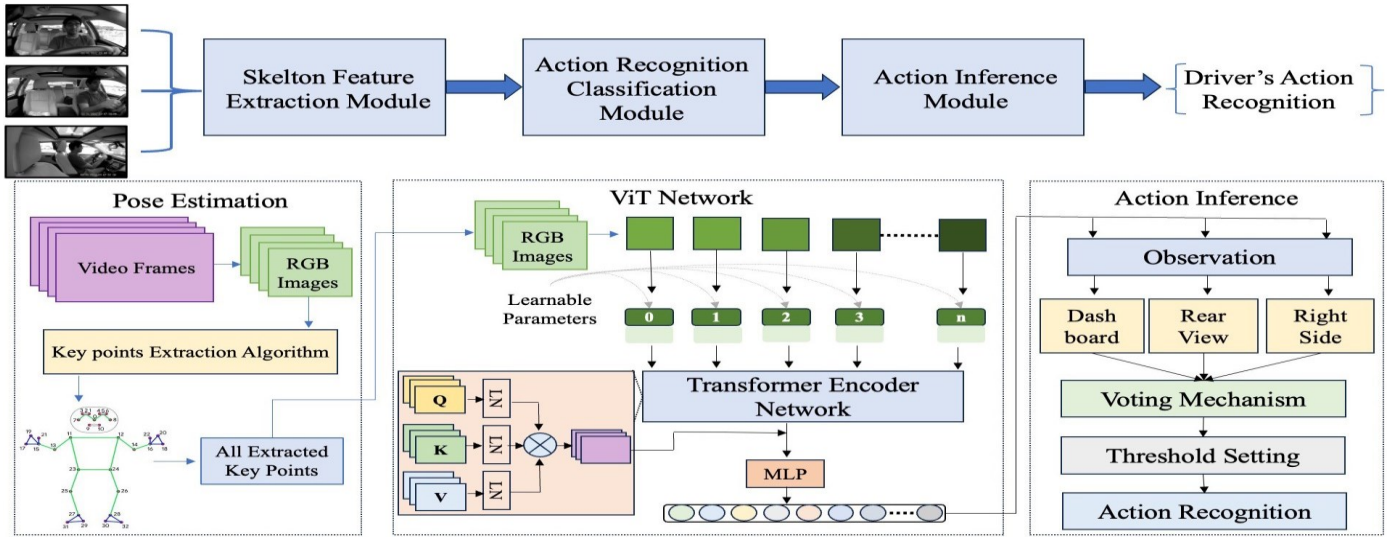

Fig. 1. Proposed PoseViNet framework adopts key points extraction through the pose estimation algorithm, ViT network for action recognition, and action inference module to optimize the output prediction.

### C. Multi-View Based Deep Learning Approaches

Multi-view deep learning empowers AI systems with a richer representation of information, improving their performance, adaptability, and robustness across diverse applications in challenging real-world scenarios. In this way, Shang et al. [50] conducted a rigorous multi-view experiment for action localization and employed multi-view synchronization across videos, and then Kinetics-700 pre-training was utilized as the feature extractor and achieved a 28.49% F1-score. Nguyen et al. [51] developed MoviNet-A0 to use data from various points of view, and some voting and post-processing techniques are applied to make a multi-view framework. Martin et al. [52] worked with several data sources to build multi-modal activity recognition. Su et al. [53] developed a classification system for multimodal and multi-view driver anomaly detection and distraction categorization using a spatial encoder with a bilateral LSTM network.

### D. Pose Estimation for Action Recognition

Pose estimation is crucial in driver behavior classification tasks, especially when dealing with significant pose variations or orientation variations. For body pose-based action detection, various state-of-the-art models based on the drivers' body poses and end-to-end architectures are analyzed [52]. For driver distraction, pose estimation of driver's head panning is targeted by Ali et al. [54]. Sharma and Kumar [55] used facial features to recognize driver fatigue using CNN architecture. Koay et al. [56] generated pose estimation images using HR-Net and ResNet [37], and thus ResNet101 and ResNet50 are used to classify the action with pose estimation. The driver's head and hand heatmaps are combined with the color image to extract pose characteristics, and keypoint classification is then used to recognize the driver's driving style [57]. Vats and Anastasiu [58] developed a technique that extracts static and movement-based properties from video's keyframes, which are then utilized to predict a series of keyframe activities. In addition to proposing fully self-attentional architectures for action recognition, Mazzia et al. [59] used a transformer encoder for 2D pose-based human action recognition.

The previous studies focus more on single-view approaches, but single-view-based approaches are susceptible to errors due to occlusions, lighting conditions, and other factors. A multi-view system can overcome these limitations by combining information from multiple cameras to provide a more complete and accurate image of the driver's behavior. Lately, researchers have presented single-directional view image or very few two-directional view image-based methods that only have limited behavior categories and compromise with real-time processing demands. Some researchers have presented multi-view video-based approaches, but these approaches cause challenges in computational complexity due to the temporal dimension. This temporal aspect introduces dependencies between frames and takes longer processing time. However, the proposed PoseViNet framework addresses these issues, using three-directional multi-view images, by providing detailed categories of driver actions through the driver's pose information and vision transformer.

## III. PROPOSED METHOD

In Fig. 1, proposed PoseViNet framework is presented. The proposed framework is based on a ViT network, which refers to getting knowledge from a pose estimation method. The proposed framework includes three steps: (i) constructing a pose estimation pipeline, (ii) a robust neural architecture using the ViT network, and (iii) an action inference module for accurate action recognition.

### A. Pose Estimation Module

This paper uses Google's open-source framework, MediaPipe Pose (MPP), to obtain 2D coordinates of human joints from images [60]. Specifically, a MediaPipe Holistic pipeline is used to get the landmarks from the face, hand, and body stance. The MPP employs a lightweight machine learning architecture known as BlazePose to process input data and extract 33 landmarks consisting of $x$, $y$, and $z$ coordinates on the human body, determining the regions of interest (ROI) within the image. The landmarks are plotted 3-dimensionally, as shown in Fig. 2 to show the complete functionality. The pose landmark within the ROI identifies postures sequentially from the ROI-cropped frame. As a result, it correctly localizes more essential points and recognizes actions appropriately. Two machine learning models, Detector and Estimator, make BlazePose. The Detector removes the human region from the input image, and the Estimator inputs a $224 \times 224$ resolution image of the discovered person and outputs the key points. Consequently, the coordinates represent the driver's position in each image frame. After processing the Detector and Estimator model, various driver's landmarks, such as eye, nose, and hand with joint positions, are defined. These extracted landmarks are connected with a line to make a complete skeleton. Different actions produce different postures; hence, these postures help extract the more reliable features from ROI. The study focuses on 25 specific landmarks from 0 to 24 consisting of $x$ and $y$ coordinates plotted on the two-dimensional image to get critical points localization. Subsequently, the dataset incorporates pose information, where these key points behave as spatial features within a neural network.

### B. ViT Network

The most recent state-of-the-art neural network, ViT, has the capacity to attain structural information from an image by employing its considerable layers in deeper networks. The ViT architecture used in this study is shown in Fig. 3.

The ViT accepts an image as input and divides the image into patches called tokens. The transformer maintains the connection between tokens. Due to the reduced pixel count within these tokens, feature extraction requires less memory and processing resources. The posture information focuses on capturing significant features from a critical portion of the image instead of the overall image. The vision transformer's input image contains the pose information generated by the pose estimation module as follows:

$$I_{ViT} = I_i \oplus L_i, \qquad I_i, L_i, I_{ViT} \in \mathbb{R}^{H \times W \times C} \qquad (1)$$

where $I_i$ represents two-dimensional image having height $H$, width $W$, and color channel $C$ taken from the $i$-th camera, where $i = 1, 2, 3$. The $L_i$ consisting of positional landmarks, $L_k^{q,r}$ for $k = 0, 1, \ldots, 24$ extracted from image $I_i$ having thickness $d$ and joint circle-radius $r$, represents two-dimensional image with height $H$, width $W$, and color channel $C$ same as those of image $I_i$. The image $L_i$ is added to the image $I_i$ using element-wise addition $\oplus$ with a condition where if $L_i$ pixel value is not zero, then replace the pixel of $I_i$ with respective pixel of $L_i$, so only positional landmarks superimpose on the image $I_i$ to create a new image $I_{ViT}$, which is the input image for vision transformer.

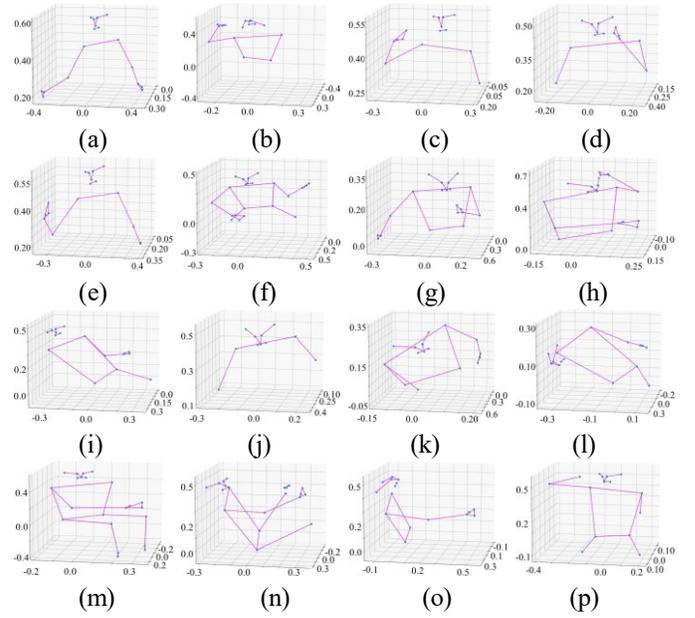

Fig. 2. Pose information extracted from images: (a) normal forward driving dashboard view; (b) drinking dashboard view; (c) phone call with right-hand dashboard view; (d) phone call with left-hand dashboard view; (e) eating dashboard view; (f) texting through right-hand rearview; (g) texting through left-hand dashboard view; (h) hair and makeup rearview; (i) reaching behind rearview; (j) adjusting the control panel rearview; (k) picking something near driver's seat rearview; (l) picking something near passenger's seat rearview; (m) talking to a passenger sitting right rearview; (n) talking to a passenger sitting back right-side view; (o) yawning right-side view; (p) hand on head dashboard view.

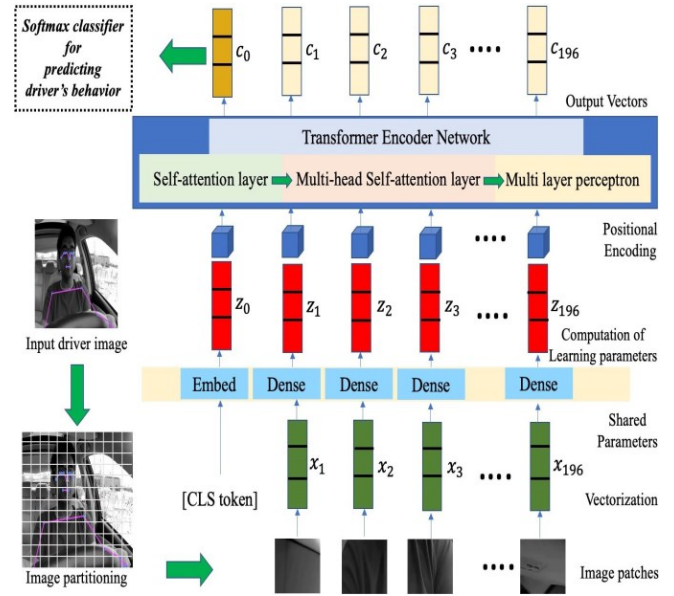

Fig. 3. Vision transformer as backbone network.

For creating set of patches $(x_n)$ having positional land-marks, a sliding window $W$ having height $p_1$ and width $p_2$ in pixels moves on image $I_{ViT}$ with stride having height $s_1$ and width $s_2$ in pixels. After every move, sliding window $W$ generates a patch having height $p_1$ and width $p_2$ in pixels. The height and width of the stride are set equal to the height and width of the patch to create non-overlapping patches while covering the whole image area. Subsequently, the vectorization function converts the patches and produces vectors. The vectors are $d_1 \times d_2 \times d_3$ dimension expressed same as $x_1, x_2, ....x_{196}$ for $d_1 \times d_2 \times d_3$ tensors patches.

The stride setting determines how the stride moves on the image every time. The $I_{ViT}$ image size is set to $224 \times 224$, and patch size and stride size are set to $16 \times 16$; consequently, the process generates 196 patches. Each patch is a tiny color image with RGB channels and three tensor order.

After this procedure, a dense layer without an activation function is used to create patch embedding through a trainable linear projection from generated set of patches $x_n$ to output $z_n$ as follows:

$$z_n = wx_n + b + W_{pos} \quad (2)$$

where $n\ (=1,2,\ldots,196)$ represents patch number. Positional encoding $W_{pos}$ is calculated using set of patches $x_n$ to augment them with positional information. Weight matrix $w$ and bias vector $b$ undergo learning through the training process.

This modification increases the capacity of the model to learn more complex representations. This $w$ and $b$ are commonly shared parameters within the dense layer. An embedding layer is appended with these dense layers that accept an additional classification token as input referred to as CLS and compute $z_0$ vector as output to the series of $z$ vectors that has the same shape as $z_1, z_2...,z_{196}$, for aggregating global image data. The output of the CLS token is fed to fully connected layers with a softmax activation function to classify extracted features and make a final prediction.

A transformer encoder network is added, which is primarily characterized by three crucial components: self-attention, multi-head self-attention, and a multilayer perceptron.

*1) Self-attention:* The attention mechanism focuses on the specific data portion. In this process, the $z_n$ is converted into a query $Q$, key $K$, and value $V$ matrices given by:

$$\begin{aligned} Q &= z_n \times w_q \\ K &= z_n \times w_k \\ V &= z_n \times w_k \end{aligned} \quad (3)$$

where $w_q$, $w_k$, and $w_v$ are learnable parameters that train during the model training process. The assignment of weights to these values is determined through the dot product operation between the query and its associated key. The calculation of the attention mechanism among different input vectors is performed as follows:

$$Attention(Q,K,V) = softmax\left(\frac{QK^T}{\sqrt{d_k}}\right) \times V \quad (4)$$

where $\sqrt{d_k}$ represents the key vector with dimension $k$, and it enhances the stability of the gradient function through appropriate normalization.

*2) Multi-Head Self-attention:* Multi-head self-attention (*mhsa*) generates a sequence of $(n+1)$ vectors from input vectors $z_0, z_1,...,z_{196}$. For each input, the multi-head function scales dot-product computations in parallel before combining the attention outputs to produce the outcome. Each head runs separately and contributes to capturing various features of the input. The equation for each head in multi-head self-attention is as follows:

$$head_j = Attention(Qw_j^Q, Kw_j^K, Vw_j^V) \quad (5)$$

where $j$ represents number of heads in multi-head and $w_j$ is trainable parameter. The *mhsa* involves calculating attention scores using query $Q$ and key matrix $K$. These scores help to weight the value matrix $V$. This process is repeated for multiple attention heads in parallel, and the results are concatenated and linearly transformed to produce the final output as follows:

$$mhsa(Q,K,V) = concat(head_1, head_2,.., head_j)w^0 \quad (6)$$

where $w^0$ matrix is the trainable parameter. The benefit is that it enables the network to incorporate sequence and location information across several representation subspaces. The *mhsa* contains a linear normalization layer *(LinNorm)* and skip connections like residual networks as:

$$\hat{c}_t = mhsa(LinNorm(c_{t-1})) + c_{t-1} \quad (7)$$

where output for layer $t$ is based on the assumption of the previous layer $t$-$1$, which is $c_{t-1}$. This is used to normalize the activations of a neural network layer to stabilize training.

*3) Multi-Layer Perceptron:* A multilayer perceptron function is used after the multi-head attention layer. To activate the layer, Gaussian Error Linear Units activation function is used as follows:

$$f(x) = x\phi(x) \quad (8)$$

where $\phi(x)$ is the standard Gaussian cumulative distribution function. Multilayer perceptron layer also contains a LinNorm as:

$$c_t = mhsa(LinNorm(\hat{c}_t)) + \hat{c}_t \quad (9)$$

where $c_t$ is the output of multi-perceptron layer and $\hat{c}_t$ is the output of previous *mhsa* layer. The transformer encoder network produces $c_0, c_1,..., c_{196}$ as output vector that performs classification. Only $c_0$ is used to make the final prediction because it has the same shape as the total number of classes, and the remaining vectors $c_1$ to $c_{196}$ are skipped. Thus, $c_0$ is fed into the softmax function to calculate a multinomial probability distribution. The layer $t$'s output is expressed as:

$$\sigma(\vec{c})_i = \frac{e^{c_i}}{\sum_{j=1}^{k} e^{c_k}} \quad (10)$$

For input vector $c_0$, $\vec{c}_i$ represents the standard exponential function where $k = 16$ defines action categories and $e^{c_{ik}}$ describes the standard exponential function applied to the output vector.

## C. Action Inference Module

The proposed framework includes an action inference module to enhance initial findings from the action recognition classification network. This design is based on the fact that different views affect diverse actions differently and that various behaviors have varying features depending on the perspective. For instance, the activity of talking to a passenger in the backseat may be hard to see from the dashboard or rearview, but it is self-evident from the right-side window perspective. Therefore, the action inference module aims to enhance the accuracy of action recognition by integrating the probability scores from different camera views for each specific action category, as shown in Fig. 4.

Training of the dashboard view, rearview, and right-side view provides a probability distribution $\sigma(\vec{c})_1$, $\sigma(\vec{c})_2$ and $\sigma(\vec{c})_3$ respectively over different action categories. The highest probabilities are elected from these distributions. The threshold is set empirically based on the probability distribution of the training data to ensure a balance between correctly identifying positive actions and minimizing any type of misclassification. These probabilities are compared with the threshold value for further processing. Selected probabilities are then averaged to obtain a single probability to get the optimum action prediction. Thus, the optimum probability determines the final prediction.

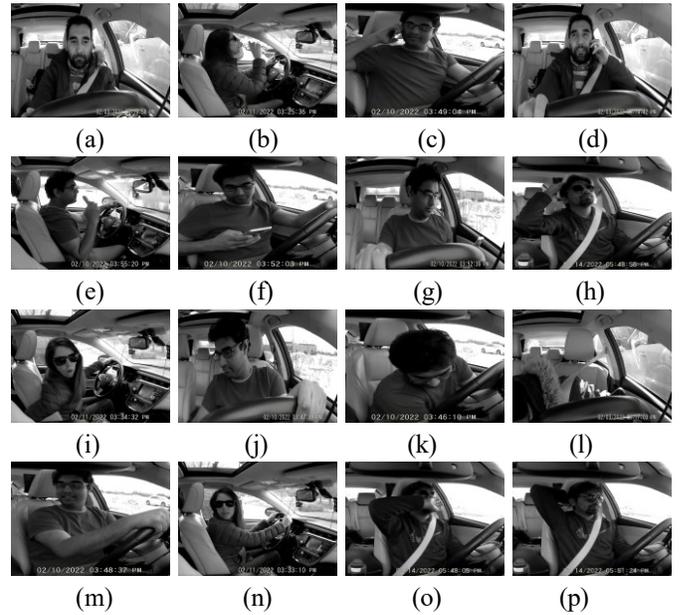

(a) (b) (c) (d)
(e) (f) (g) (h)
(i) (j) (k) (l)
(m) (n) (o) (p)

Fig. 5. Sample images from the dataset: (a) normal forward driving dashboard view; (b) drinking right-side view; (c) phone call with right-hand rearview; (d) phone call with left-hand dashboard view; (e) eating right-side view; (f) texting through right-hand rearview; (g) texting through left-hand dashboard view; (h) hair and makeup rearview; (i) reaching behind right-side view; (j) adjusting the control panel dashboard view; (k) picking something near driver's seat rearview; (l) picking something near passenger's seat dashboard view; (m) talking to passenger sitting right rearview; (n) talking to passenger sitting back right-side view; (o) yawning rearview; (p) hand on head rearview.

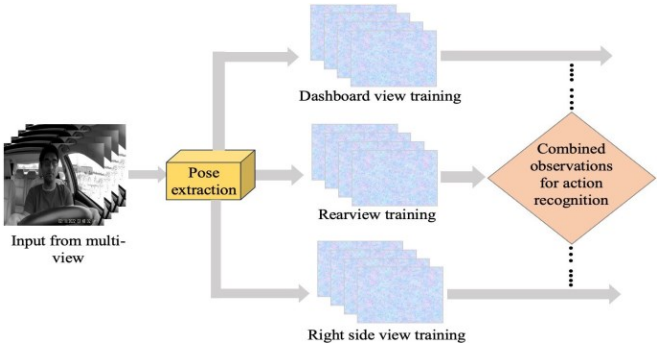

Fig. 4. Action inference module takes input from multiple views for accurate decision-making in action recognition.

## IV. EXPERIMENTS

### A. Dataset

We evaluate the proposed framework using the SynDD1 dataset [61] widely used to study distracted driving. The dataset provides information on realistic distracted driving behavior. Videos of distracted driving of the SynDD1 dataset are captured by cabin cameras within the car. A total of 18 individuals engaged in various driving activities inside a car participated to create the dataset. Video recordings are made from three distinct viewpoints: dashboard camera, rearview camera, and right-side window camera. Each distracted driving video in the dataset has a frame rate of 30 FPS and an RGB resolution of (1080,1920,3) for each frame. Out of 18 driver behaviors, 16 are selected for the experiment. Driver's singing and dancing behavior is excluded from the study because it is difficult to identify their visual identification from image data. Driver's different actions are labeled to evaluate the framework, as mentioned in Table I. Samples of distracted driving action frames from various views are shown in Fig. 5.

### B. Experiment Setting

Specific training configurations used in the experiments are listed in Table II. The proposed framework is trained by using Tensorflow 2.12.0 and Keras 2.12.0 packages in the deep-learning framework employing Anaconda Navigator 2.4.2 in a Python 3.11.4 environment. The images extracted from the various videos in the dataset are divided into three sets in the 0.7:0.15:0.15 proportional ratios: training set, validation set, and testing set. One hot encoding is performed on respective action labels for better model presentation. Different driver action images are used in the experiments to analyze the performance of the AdamW optimizer. The AdamW optimizer experiences minimal training loss while training models, enabling it to learn the weights for a generalized trained model with an unseen testing set. The AdamW optimizer performs better handling of weight decay as a regularization, which is used to prevent overfitting of the model by penalizing large weight values. In the training phase, the optimizer provides independent optimization of the learning rate and weight decay. Therefore, changing the weight decay value is not necessary when optimizing the learning rate. The proposed framework with this characteristic makes it more broadly applicable. The learning rate is initially set to 0.001 throughout the model optimization phase. After that, at timestamp t, each batch of the driver's action images used for training updates the model's weights and biases during optimization and returns respective gradients. Consequently, the AdamW optimizer improves regularization by decoupling the weight decay from the updated gradient. This procedure is carried out for 50 epochs to update the weights and reduce the loss.

TABLE I
DRIVER BEHAVIOUR FOR DISTRACTED DRIVING IN THE SYNDD1 DATASET

| Label | Driver's action while driving the car |
|---|---|
| C1 | driver is driving normally and forward |
| C2 | driver is drinking from water bottle |
| C3 | driver is on a phone call with right hand |
| C4 | driver is on a phone call with left hand |
| C5 | driver is eating |
| C6 | driver is texting through right hand |
| C7 | driver is texting through left hand |
| C8 | driver is setting her or his hair and makeup |
| C9 | driver is reaching behind towards backseat |
| C10 | driver is adjusting the control panel |
| C11 | driver is picking up something near the driver's seat floor |
| C12 | driver is picking up something near the passenger's seat floor |
| C13 | driver is talking to the passenger sitting in the right seat |
| C14 | driver is talking to the passengers sitting in the backseat |
| C15 | driver is yawning |
| C16 | driver's hand is on head |

TABLE II
HYPERPARAMETERS AND TRAINING ENVIRONMENT

| Parameter | Configuration details |
| --- | --- |
| Platform | Intel i7-6700 + NVIDIA GeForce GTX 1080 Ti + 32GB RAM |
| Operating System | Windows 10 Pro |
| Framework | TensorFlow 2.12.0 |
| Optimizer | AdamW ($\beta_1$ =0.9, $\beta_2$ =0.999) |
| Learning Rate | 0.001 |
| Weight Decay | 0.0001 |
| Batch Size | 32 |
| Patch Size | 16 |
| Head Numbers | 4 |
| Epochs | 50 |
| Activation Function | Gaussian Error Linear Units, Softmax |
| Loss Function | Categorical Crossentropy Loss |
| Dropout | 25%, 50% |

### C. Evaluation Metrics

Performance metrics such as accuracy, precision rate, recall rate, specificity, false positive rate (FPR), and the F1-score are used to evaluate the reliability of the framework. A total number of true positives, true negatives, false positives, and false negatives are calculated from the dashboard view, rearview, and right-side view test sets to compute these performance metrics.

## V. RESULTS AND DISCUSSION

The PoseViNet framework containing the ViT architecture as well as pose estimation module and action inference module has been designed and proposed for the multi-view driver action recognition. Experiments with the comprehensive SynDD1 dataset are conducted with specific experiment settings described in subsections IV.A and IV.B.

The number of frames for the "phone call with right hand" class is much larger than the number of frames for the "hair and makeup" and "reaching behind" class. Thus, we randomly choose images from each category to balance the data and ensure that each category represents almost the same number of frames. Table III presents the data distribution, which displays selected frames of each class. A total of 72,720 frames combining each 24,240 ones from the dashboard, rearview, and right-side views are used to perform the experiments using the proposed framework. In the training of the dashboard view, 16,968 frames are kept as the training set, 3,636 frames as the validation set, and 3,636 frames as the testing test. The exact configuration is also applied for the training with rearview and right-side view frames.

### A. Experimental Results with Deep Learning

The proposed framework is trained and evaluated using a specific set of hyperparameters, and all data samples are pre-processed and resized following the input dimensions of the training of the ViT architecture. The performance of the AdamW optimizer is evaluated by using images of different driver actions in the experiments.

The self-attention layer, multi-head self-attention layer, and multilayer perceptron with dropouts, and fully-connected layers are added to the design of the ViT network. In this experiment, four multi-head, multilayer perceptron layers are added and normalized using the keras layer normalization method. Pose information of various driver actions is extracted from the categories and injected into ViT network to extract more prominent features, which leads to better training performance. The ViT network is trained with various architectural settings with and without pose information.

The deep neural network with the best performance is determined by using comparison criteria that includes testing and validation accuracy. The average validation and testing accuracies of the ViT network are 97.55% and 90.92%, respectively, when pose information is fed into the ViT network for training. The ViT network achieves almost similar training accuracy when it is trained alone; however, the model is unable to distinguish patterns like ViT network with pose information architecture.

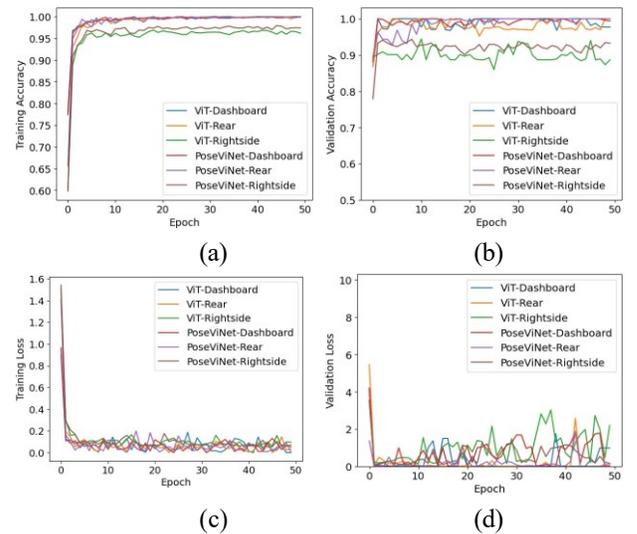

Fig. 6. Visualized representation of curves obtained from network training: (a) training accuracy; (b) validation accuracy; (c) training loss; (d) validation loss.

TABLE III
DATA DISTRIBUTION FOR TRAINING PROPOSED FRAMEWORK: POSEVINET

| Class | Dashboard view | | | Rearview | | | Right-side view | | |
|---|---|---|---|---|---|---|---|---|---|
| | Training | Validation | Testing | Training | Validation | Testing | Training | Validation | Testing |
| Normal | 1092 | 234 | 234 | 1092 | 234 | 234 | 1092 | 234 | 234 |
| Drinking | 1050 | 225 | 225 | 1050 | 225 | 225 | 1050 | 225 | 225 |
| Phone call (right-hand) | 1064 | 228 | 228 | 1064 | 228 | 228 | 1064 | 228 | 228 |
| Phone call (left-hand) | 1050 | 225 | 225 | 1050 | 225 | 225 | 1050 | 225 | 225 |
| Eating | 1036 | 222 | 222 | 1036 | 222 | 222 | 1036 | 222 | 222 |
| Texting (right-hand) | 1064 | 228 | 228 | 1064 | 228 | 228 | 1064 | 228 | 228 |
| Texting (left-hand) | 1092 | 234 | 234 | 1092 | 234 | 234 | 1092 | 234 | 234 |
| Hair and makeup | 1036 | 222 | 222 | 1036 | 222 | 222 | 1036 | 222 | 222 |
| Reaching behind | 1050 | 225 | 225 | 1050 | 225 | 225 | 1050 | 225 | 225 |
| Adjusting control panel | 1064 | 228 | 228 | 1064 | 228 | 228 | 1064 | 228 | 228 |
| Picking near driver's seat | 1036 | 222 | 222 | 1036 | 222 | 222 | 1036 | 222 | 222 |
| Picking near passenger's seat | 1050 | 225 | 225 | 1050 | 225 | 225 | 1050 | 225 | 225 |
| Talking to passenger at right | 1036 | 222 | 222 | 1036 | 222 | 222 | 1036 | 222 | 222 |
| Talking to passenger at backside | 1092 | 234 | 234 | 1092 | 234 | 234 | 1092 | 234 | 234 |
| Yawning | 1064 | 228 | 228 | 1064 | 228 | 228 | 1064 | 228 | 228 |
| Hand on head | 1092 | 234 | 234 | 1092 | 234 | 234 | 1092 | 234 | 234 |
| **Total** | **16968** | **3636** | **3636** | **16968** | **3636** | **3636** | **16968** | **3636** | **3636** |

TABLE IV
TRAINING PERFORMANCE OF PROPOSED FRAMEWORK WITH DIFFERENT ARCHITECTURAL SETTINGS

| | Training Acc. | Validation Acc. | Testing Acc. |
|---|---|---|---|
| ViT | 98.68% | 95.45% | 87.42% |
| Pose + ViT + Action Inference | 99.09% | 97.55% | 90.92% |

Table IV provides the results, such as training and validation accuracy, including testing accuracy with unseen driver action images obtained from the experiments on different architectural arrangements. Compared to the ViT architecture, the PoseViNet obtains a noticeable accuracy. The accuracy and loss curves for training and validation obtained from various architectural configurations with optimizer and learning rate settings are presented in Fig.6.

### B. Comparison with state-of-the-art models

The performance of the PoseViNet is compared with state-of-the-art deep learning models to analyze the prediction assessment with unseen driver action data. We compare the proposed framework with three different state-of-the-art networks: InceptionNet-V4 [62], DenseNet [48], and VGG-19 [35], all of which are known for reliable in classification and extensively applied in research with noteworthy performance. These networks are trained using the same techniques outlined in subsection IV.B for fair comparisons. Fig.7 illustrates the accuracy and loss curves for training and validation obtained from the specific settings of several state-of-the-art models. The PoseViNet framework attained a noticeable accuracy curve compared with other models. Table V represents the results of the experiment with the state-of-the-art models and lists the training and validation accuracy of each model. The model with the best training accuracy is InceptionNet-V4; however, it is not as prominent at pattern discrimination as PoseViNet because the model having a high validation accuracy indicates good generalization to unseen driver behavior data without overfitting to the training data. The best model weights across all networks are saved for model testing. The proposed framework outperforms other state-of-the-art ones in certain optimizer and learning rate combinations.

Table VI provides details of the model's size and inference times for unseen images. PoseViNet demonstrated a notable inference time of 15.16 seconds on the testing set, which is significantly less than other approaches listed in the table. The testing is assessed with 34 batches, and the proposed framework achieved an average processing speed of 399 milliseconds per step. The trained model size indicates that the computational cost of the proposed framework is very low, and the inference time indicates that the model is faster than other approaches. Table VII provides a complete model analysis of the proposed framework regarding precision, accuracy, recall, F1-score, specificity, and FPR.

TABLE V
TRAINING OF DIFFERENT DEEP LEARNING NETWORKS WITH SPECIFIED HYPERPARAMETERS

| | Training Acc. | Validation Acc. | Testing Acc. |
|---|---|---|---|
| InceptionNet-V4 | 99.19% | 88.91% | 82.43% |
| DenseNet | 94.86% | 82.16% | 80.18% |
| VGG19 | 93.88% | 79.98% | 74.25% |
| PoseViNet | 99.09% | 97.55% | 90.92% |

TABLE VI
ANALYSIS OF COMPUTATIONAL REQUIREMENTS AND REAL-TIME PERFORMANCE OF TRAINED NETWORKS

| Network | Trained Weight Size | Inference Time | Avg Time per Step |
|---|---|---|---|
| InceptionNet-V4 | 93.3 MB | 27.73s | 781ms |
| DenseNet | 88.9 MB | 99.46s | 3s |
| VGG19 | 92.5 MB | 127.25s | 4s |
| PoseViNet | 36.7 MB | 15.96s | 399ms |

Confusion matrix For PoseViNet on unseen driver action images is presented in Fig. 8. The proposed PoseViNet

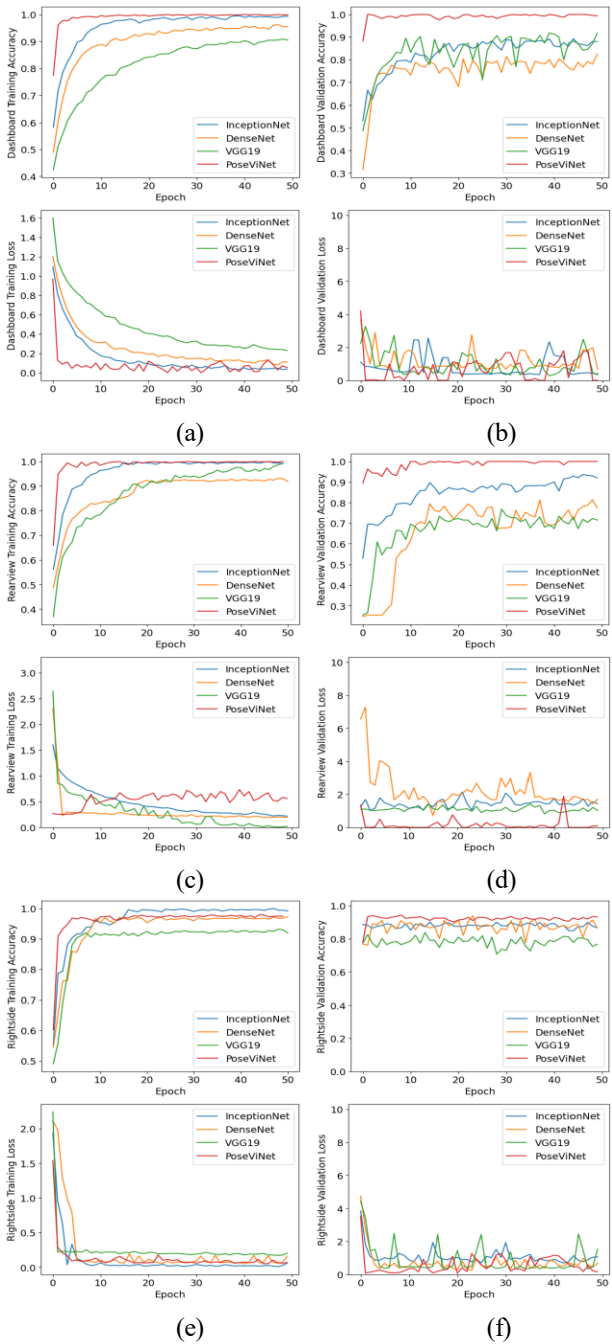

Fig. 7. Visualized representation of curves obtained from different deep learning network training: (a) dashboard training accuracy; dashboard training loss; (b) dashboard validation accuracy; dashboard validation loss; (c) rearview training accuracy; rearview training loss; (d) rearview validation accuracy; rearview validation loss; (e) right-side training accuracy; right-side training loss; (f) right-side validation accuracy; right-side validation loss.

achieves 90.92% accuracy on a large dataset.

## C. Comparison with other baseline methods

Recent image-based approaches used only one view dataset, mainly data from a rearview angle for 10 behaviors of driver. The dataset SFD3 [63] has been widely used in baseline methods by researchers to validate their experiments. Mittal

TABLE VII
PRECISION, RECALL, F1-SCORE, SPECIFICITY, FPR, AND ACCURACY FOR POSEVINET ON UNSEEN DRIVER ACTION IMAGES BELONGING TO TESTING SET

| Class | Precision | Recall | F1-Score | Specificity | FPR | Accuracy |
|---|---|---|---|---|---|---|
| C1 | 0.84 | 1.00 | 0.91 | 0.99 | 0.01 | 0.99 |
| C2 | 0.43 | 0.95 | 0.59 | 0.92 | 0.08 | 0.92 |
| C3 | 0.71 | 0.91 | 0.80 | 0.98 | 0.02 | 0.97 |
| C4 | 0.25 | 0.89 | 0.39 | 0.83 | 0.17 | 0.83 |
| C5 | 0.47 | 0.89 | 0.62 | 0.94 | 0.06 | 0.93 |
| C6 | 0.62 | 0.94 | 0.75 | 0.96 | 0.04 | 0.96 |
| C7 | 0.21 | 0.89 | 0.34 | 0.77 | 0.23 | 0.78 |
| C8 | 0.44 | 0.97 | 0.60 | 0.92 | 0.08 | 0.92 |
| C9 | 0.44 | 0.95 | 0.61 | 0.92 | 0.08 | 0.92 |
| C10 | 0.43 | 0.90 | 0.58 | 0.92 | 0.08 | 0.92 |
| C11 | 0.34 | 0.90 | 0.50 | 0.89 | 0.11 | 0.89 |
| C12 | 0.28 | 0.87 | 0.42 | 0.85 | 0.15 | 0.85 |
| C13 | 0.58 | 0.95 | 0.72 | 0.95 | 0.05 | 0.95 |
| C14 | 0.39 | 0.93 | 0.54 | 0.90 | 0.10 | 0.90 |
| C15 | 0.30 | 0.92 | 0.45 | 0.85 | 0.15 | 0.86 |
| C16 | 0.58 | 1.00 | 0.73 | 0.95 | 0.05 | 0.95 |
| Average | 0.45 | 0.92 | 0.59 | 0.90 | 0.09 | 0.90 |

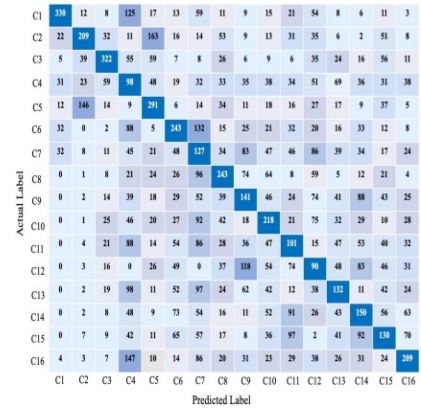

Fig. 8. Confusion matrix for PoseViNet on unseen driver action images.

and Verma [45] achieved 99.88% accuracy with the SFD3, whereas Liu et al. [46] achieved 99.87% accuracy. Dhakate and Dash [43] performed experiments with the SFD3 and achieved 97.00% accuracy. Aljohani [47] obtained 99.80% and Ma and Wang [49] achieved 92.51%. To make a fair comparison with these baseline results, the PoseViNet is trained with the SFD3, and achieved 99.96% accuracy. The proposed framework is trained on SFD3 to evaluate the performance in different driving environment settings. This variety ensures that the model can generalize well to diverse real-time driving conditions. Table VIII shows the comparison with different works. As seen in the table. proposed framework slightly or significantly outperforms other baseline methods

An essential facet of road safety, yet a common real-world challenge, is establishing dependable action recognition that accommodates the diverse viewpoints of drivers. The integration of pose estimation into the ViT architecture has the potential to enhance their performance in multi-view driver distraction recognition. Pose estimation offers valuable context regarding the driver's body posture, providing a comprehensive understanding of their behavior. By incorporating pose infor-mation, the PoseViNet framework can extract more relevant features from varying driver poses and viewpoints, resulting in more accurate and reliable distraction detection.

TABLE VIII
COMPARISON OF PROPOSED FRAMEWORK WITH OTHER BASELINE METHODS ON SFD3 DATASET

| | View | No. of Classes | Accuracy |
|---|---|---|---|
| Mittal and verma (2023) [45] | 1 | 10 | 99.88% |
| Liu et al. (2023) [46] | 1 | 10 | 99.87% |
| Dhakate and Dash (2020) [43] | 1 | 10 | 97.00% |
| Aljohani (2023) [47] | 1 | 10 | 99.80% |
| Ma and Wang (2023) [49] | 1 | 10 | 93.59% |
| **Proposed framework- PoseViNet** | **1** | **10** | **99.96%** |

Consequently, combining pose estimation with vision transformers presents an opportunity to significantly enhance the accuracy and robustness of driver distraction detection systems, particularly in situations involving multiple viewing angles. This integrated approach holds promise for advancing road safety by improving our comprehension of and response to driver distractions.

## VI. CONCLUSION

One of the primary factors leading to a significant increase in traffic accidents is driver distraction. This problem can be solved by implementing a vehicle-mounted automated system that continually checks the drivers' concentration and alerts if any distractions occur. It's evident that adjusting the camera angle away from the dashboard can enhance our ability to identify specific actions. For instance, actions such as reaching behind or engaging with passengers in the backseat may exhibit similar visual characteristics from the dashboard. Hence, if we merge the features extracted from the images obtained by the dashboard, rearview and right-side cameras, we can derive more valuable insights. From this viewpoint, this paper proposes a framework, namely PoseViNet, combining multi-view pose estimation with a vision transformer. The driver's posture information is injected into the transformer so that the self-attention mechanism and the multi-layer neural network architecture could extract more enhanced information from the relevant portion of the image, making the proposed framework more accurate. Pose information makes the PoseViNet robust to variations in lighting conditions and occlusions, and it efficiently captures the spatial relationships between different parts of the body and improves the generalization performance and interpretability of the PoseViNet. With one view image dataset SFD3, our approach achieves performance improved over other baseline methods. With muti-view image dataset SynDD1, 97.55% validation accuracy and 90.92% testing accuracy on unseen driver's action images are obtained respectively.

As a future work, the proposed framework can be targeted to recognize drivers' actions with their emotions. Furthermore, sensor data from onboard diagnostics, such as speed, engine revolutions, fuel usage, and more, can be fed as input to the model, and a multi-modal system can be developed with this proposed framework to recognize the driver's identity and driver's distracted actions with their driving behavior. Along with this, the output of the proposed framework can be transferred to the vehicle owner to make them aware of the driver's behavior while driving.

## ACKNOWLEDGMENTS

This work is supported by the Institute for Information communications Technology Promotion (IITP) grant funded by the Korean government (MSIT) (No.2020-0-00440, Development of Artificial Intelligence Technology that continuously improves itself as the situation changes in the real world).

AUTHOR et al.: PREPARATION OF PAPERS FOR IEEE TRANSACTIONS AND JOURNALS (FEBRUARY 2017) 11[19] K. Ramana et al., "A vision transformer approach for traffic congestion prediction in urban areas," IEEE Transactions on Intelligent Transportation Systems, vol. 24, no. 4, pp. 3922–3934, 2023. doi:10.1109/tits.2022.3233801.

[20] X. Zhu et al., "Deformable detr: Deformable Transformers for end-to-end object detection," arXiv.org, https://arxiv.org/abs/2010.04159 (accessed Jul. 12, 2023).

[21] M. Kumar, D. Weissenborn, and N. Kalchbrenner, "Colorization transformer," arXiv.org, https://arxiv.org/abs/2102.04432 (accessed Jul. 12, 2023).

[22] Rethinking semantic segmentation from a sequence-to-sequence ..., https://ieeexplore.ieee.org/document/9578646 (accessed Jul. 12, 2023).

[23] A. Arnab et al., "Vivit: A video vision transformer," arXiv.org, https://arxiv.org/abs/2103.15691 (accessed Jul. 12, 2023).

[24] H. Chen et al., "Pre-Trained Image Processing Transformer," arXiv.org, https://arxiv.org/abs/2012.00364 (accessed Jul. 12, 2023).

[25] M. Naseer et al., "Intriguing properties of vision transformers," arXiv.org, https://arxiv.org/abs/2105.10497 (accessed Jul. 12, 2023).

[26] Q.-V. Lai-Dang, J. Lee, B. Park, and D. Har, "Sensor fusion by spatial encoding for autonomous driving," arXiv.org, https://arxiv.org/abs/2308.10707 (accessed Oct. 6, 2023).

[27] P. K. Rajendran, S. Mishra, L. F. Vecchietti, and D. Har, "RelMobNet: End-to-end relative camera pose estimation using a robust two-stage training," Lecture Notes in Computer Science, pp. 238–252, 2023. doi:10.1007/978-3-031-25075-0_18

[28] X. Zhang, N. Zheng, F. Wang, and Y. He, "Visual recognition of driver hand-held cell phone use based on hidden CRF," Proceedings of 2011 IEEE International Conference on Vehicular Electronics and Safety, 2011. doi:10.1109/icves.2011.5983823.

[29] K. Seshadri, F. Juefei-Xu, D. K. Pal, M. Savvides, and C. P. Thor, "Driver cell phone usage detection on Strategic Highway Research Program (SHRP2) face view videos," 2015 IEEE Conference on Computer Vision and Pattern Recognition Workshops (CVPRW), 2015. doi:10.1109/cvprw.2015.7301397.

[30] C. Craye and F. Karray, "Driver distraction detection and recognition using RGB-D sensor," arXiv.org, https://arxiv.org/abs/1502.00250 (accessed Jul. 27, 2023).

[31] C. H. Zhao, B. L. Zhang, J. He, and J. Lian, "Recognition of driving postures by Contourlet transform and random forests," IET Intelligent Transport Systems, vol. 6, no. 2, p. 161, 2012. doi:10.1049/iet-its.2011.0116.

[32] I. Gupta, N. Garg, A. Aggarwal, N. Nepalia, and B. Verma, "Real-time driver's drowsiness monitoring based on dynamically varying threshold," 2018 Eleventh International Conference on Contemporary Computing (IC3), 2018. doi:10.1109/ic3.2018.8530651.

[33] C. Yan, F. Coenen, and B. Zhang, "Driving posture recognition by convolutional neural networks," IET Computer Vision, vol. 10, no. 2, pp. 103–114, 2016. doi:10.1049/iet-cvi.2015.0175.

[34] A. Koesdwiady, S. M. Bedawi, C. Ou, and F. Karray, "End-to-end deep learning for driver distraction recognition," Lecture Notes in Computer Science, pp. 11–18, 2017. doi:10.1007/978-3-319-59876-5_2.

[35] K. Simonyan and A. Zisserman, "Very deep convolutional networks for large-scale image recognition," arXiv.org, https://arxiv.org/abs/1409.1556 (accessed Jul. 27, 2023).

[36] M. D. Hssayeni, S. Saxena, R. Ptucha, and A. Savakis, "Distracted driver detection: Deep learning vs handcrafted features," Electronic Imaging, vol. 29, no. 10, pp. 20–26, 2017. doi:10.2352/issn.2470-1173.2017.10.imawm-162.

[37] K. He, X. Zhang, S. Ren, and J. Sun, "Deep residual learning for image recognition," 2016 IEEE Conference on Computer Vision and Pattern Recognition (CVPR), 2016. doi:10.1109/cvpr.2016.90.

[38] A. Krizhevsky, I. Sutskever, and G. E. Hinton, "ImageNet classification with deep convolutional Neural Networks," Communications of the ACM, vol. 60, no. 6, pp. 84–90, 2017. doi:10.1145/3065386.

[39] C. Szegedy, V. Vanhoucke, S. Ioffe, J. Shlens, and Z. Wojna, "Rethinking the inception architecture for computer vision," 2016 IEEE Conference on Computer Vision and Pattern Recognition (CVPR), 2016. doi:10.1109/cvpr.2016.308.

[40] Y. Abouelnaga, H. M. Eraqi, and M. N. Moustafa, "Real-time distracted driver posture classification," arXiv.org, https://arxiv.org/abs/1706.09498 (accessed Jul. 27, 2023).

[41] H. M. Eraqi, Y. Abouelnaga, M. H. Saad, and M. N. Moustafa, "Driver distraction identification with an ensemble of Convolutional Neural Networks," Journal of Advanced Transportation, vol. 2019, pp. 1–12, 2019. doi:10.1155/2019/4125865.

[42] F. Chollet, "Xception: Deep learning with depthwise separable convolutions," 2017 IEEE Conference on Computer Vision and Pattern Recognition (CVPR), 2017. doi:10.1109/cvpr.2017.195.

[43] K. R. Dhakate and R. Dash, "Distracted driver detection using stacking ensemble," 2020 IEEE International Students' Conference on Electrical, Electronics and Computer Science (SCEECS), 2020. doi:10.1109/sceecs48394.2020.184.

[44] M. Alotaibi and B. Alotaibi, "Distracted driver classification using Deep Learning," Signal, Image and Video Processing, vol. 14, no. 3, pp. 617–624, 2019. doi:10.1007/s11760-019-01589-z.

[45] H. Mittal and B. Verma, "Cat-capsnet: A convolutional and attention based capsule network to detect the driver's distraction," IEEE Transactions on Intelligent Transportation Systems, pp. 1–10, 2023. doi:10.1109/tits.2023.3266113.

[46] D. Liu, T. Yamasaki, Y. Wang, K. Mase, and J. Kato, "Toward extremely lightweight distracted driver recognition with distillation-based neural architecture search and knowledge transfer," IEEE Transactions on Intelligent Transportation Systems, vol. 24, no. 1, pp. 764–777, 2023. doi:10.1109/tits.2022.3217342.

[47] Abeer. A. Aljohani, "Real-time driver distraction recognition: A hybrid genetic deep network based approach," Alexandria Engineering Journal, vol. 66, pp. 377–389, 2023. doi:10.1016/j.aej.2022.12.009.

[48] G. Huang, Z. Liu, L. Van Der Maaten, and K. Q. Weinberger, "Densely connected Convolutional Networks," 2017 IEEE Conference on Computer Vision and Pattern Recognition (CVPR), 2017. doi:10.1109/cvpr.2017.243.

[49] Y. Ma and Z. Wang, "VIT-DD: Multi-task vision transformer for semi-supervised driver Distraction detection," arXiv.org, https://arxiv.org/abs/2209.09178 (accessed Sep. 5, 2023).

[50] J. Shang, J. Cao, K. Li, K. Tian, and X. Zhang, "MVP: Robust multi-view practice for driving action localization," 2022 IEEE 5th International Conference on Information Systems and Computer Aided Education (ICISCAE), 2022. doi:10.1109/iciscae55891.2022.9927603.

[51] H.-Q. Nguyen et al., "End-to-end deep learning-based framework for Driver Action Recognition," 2022 International Conference on Multimedia Analysis and Pattern Recognition (MAPR), 2022. doi:10.1109/mapr56351.2022.9924944.

[52] M. Martin et al., "Drive&Act: A multi-modal dataset for fine-grained driver behavior recognition in Autonomous Vehicles," 2019 IEEE/CVF International Conference on Computer Vision (ICCV), 2019. doi:10.1109/iccv.2019.00289.

[53] L. Su, C. Sun, D. Cao, and A. Khajepour, "Efficient driver anomaly detection via conditional temporal proposal and Classification Network," IEEE Transactions on Computational Social Systems, vol. 10, no. 2, pp. 736–745, 2023. doi:10.1109/tcss.2022.3158480.

[54] S. F. Ali, A. S. Aslam, M. J. Awan, A. Yasin, and R. Damaševičius, "Pose estimation of driver's head panning based on interpolation and motion vectors under a boosting framework," Applied Sciences, vol. 11, no. 24, p. 11600, 2021. doi:10.3390/app112411600.

[55] S. Sharma and V. Kumar, "Distracted driver detection using learning representations," Multimedia Tools and Applications, vol. 82, no. 15, pp. 22777–22794, 2023. doi:10.1007/s11042-023-14635-3.

[56] H. V. Koay, J. H. Chuah, C.-O. Chow, Y.-L. Chang, and B. Rudrusamy, "Optimally-weighted image-pose approach (OWIPA) for distracted driver detection and classification," Sensors, vol. 21, no. 14, p. 4837, 2021. doi:10.3390/s21144837.

[57] M. Lu, Y. Hu, and X. Lu, "Pose-guided model for driving behavior recognition using keypoint action learning," Signal Processing: Image Communication, vol. 100, p. 116513, 2022. doi:10.1016/j.image.2021.116513.

[58] A. Vats and D. C. Anastasiu, "Key point-based driver activity recognition," 2022 IEEE/CVF Conference on Computer Vision and Pattern Recognition Workshops (CVPRW), 2022. doi:10.1109/cvprw56347.2022.00370.

[59] V. Mazzia, S. Angarano, F. Salvetti, F. Angelini, and M. Chiaberge, "Action transformer: A self-attention model for short-time pose-based human action recognition," Pattern Recognition, vol. 124, p. 108487, 2022. doi:10.1016/j.patcog.2021.108487.

[60] "MediaPipe framework: google for developers," Google, https://developers.google.com/mediapipe/solutions/vision/pose_landmarker (accessed Aug. 14, 2023).

[61] M. S. Rahman et al., "Synthetic distracted driving (syndd1) dataset for analyzing distracted behaviors and various gaze zones of a driver," Data in Brief, vol. 46, p. 108793, 2023. doi:10.1016/j.dib.2022.108793.

[62] C. Szegedy, S. Ioffe, V. Vanhoucke, and A. Alemi, "Inception-V4, inception-resnet and the impact of residual connections on learning,"